%% file: iclr2023_conference.tex
\documentclass{article} % For LaTeX2e
\usepackage{iclr2023_conference,times}

\input{math_commands.tex}

\usepackage{hyperref}
\usepackage{url}
\usepackage{graphicx}
\usepackage{siunitx}
\usepackage{textcomp}

\title{MatKB: Semantic Search for Polycrystalline Materials Synthesis Procedures}

\author{Antiquus S.~Hippocampus, Natalia Cerebro \& Amelie P. Amygdale \thanks{ Use footnote for providing further information
about author (webpage, alternative address)---\emph{not} for acknowledging
funding agencies.  Funding acknowledgements go at the end of the paper.} \\
Department of Computer Science\\
Cranberry-Lemon University\\
Pittsburgh, PA 15213, USA \\
\texttt{\{hippo,brain,jen\}@cs.cranberry-lemon.edu} \\
\And
Ji Q. Ren \& Yevgeny LeNet \\
Department of Computational Neuroscience \\
University of the Witwatersrand \\
Joburg, South Africa \\
\texttt{\{robot,net\}@wits.ac.za} \\
\AND
Coauthor \\
Affiliation \\
Address \\
\texttt{email}
}

\begin{document}

\maketitle

\begin{abstract}
In this paper, we present a novel approach to knowledge extraction and retrieval using Natural Language Processing (NLP) techniques for material science. Our goal is to automatically mine structured knowledge from millions of research articles in the field of polycrystalline materials and make it easily accessible to the broader community. The proposed method leverages NLP techniques such as entity recognition and document classification to extract relevant information and build an extensive knowledge base, from a collection of 9.5 Million publications. The resulting knowledge base is integrated into a search engine, which enables users to search for information about specific materials, properties, and experiments with greater precision than traditional search engines like Google. We hope our results can enable material scientists quickly locate desired experimental procedures, compare their differences, and even inspire them to design new experiments. Our website will be available at Github \footnote{https://github.com/Xianjun-Yang/PcMSP.git} soon.
\end{abstract}

\section{Introduction}

Materials science is a rapidly growing and evolving field, with discoveries and innovations always being made. As the field grows, so does the amount of published research, making it increasingly challenging for researchers to keep up with the latest developments and find the information they need. This is especially true for researchers working in specialized areas, where the sheer volume of research can make it difficult to find relevant information. Therefore, there has been growing interest in applying machine learning for automatically extracting information from tons of publications \cite{kim2017materials, olivetti2020data}.

Traditionally, researchers have relied on search engines like Google to find information. While these search engines are powerful and widely used, they can be limited in their ability to search within specific fields, such as materials science. Additionally, they often return many irrelevant results, making it time-consuming to sort through the results and find the information one needs.

To address these challenges, this paper presents a new approach to knowledge extraction and retrieval using NLP techniques. Our approach leverages the advances in NLP to automatically extract relevant information from research articles, such as materials, properties, and experiments, and build a large knowledge base. This knowledge base is then integrated into a search engine that allows users to search for information about specific materials and experiments with greater precision and speed than traditional search engines.

Recently, there has been a released corpus PcMSP \cite{yang2022pcmsp} for entities and relations extraction from polycrystalline materials synthesis procedure. We utilize their data to build our search engine as a first step. We leave the extension to the whole materials domain for future work. 

In general, from a collection of 4.9M and 4.6M publications in physics and material science domain in S2ORC \cite{lo-etal-2020-s2orc}, we retrieve 5,846 relevant articles. Based on this, we extract 269,808 desired entities for constructing our semantic search platform MatKB. Compared with the human expert-curated commercial application like Reaxsys provided by Elsevier, we will make our platform freely available to the public.

\section{Related Work}
The application of Natural Language Processing (NLP) techniques in materials science has gained significant attention in recent years. The main objective of using NLP in materials science is to extract information from unstructured text sources such as scientific articles, patents, and technical reports. This information can be used for various purposes such as knowledge discovery, material design, and performance optimization.

One of the earliest studies on NLP for materials procedures extraction was performed by \cite{kim2017materials}, who used NLP techniques to extract materials processing information from the literature. They proposed a system that used rule-based and machine learning-based methods to identify and extract materials processing information and make predictions based on it. Similar work has also been reported in \cite{jensen2019machine, kim2017machine, he2020similarity}.

In conclusion, using NLP techniques for materials procedure extraction has shown promising results and has the potential to revolutionize the way information is extracted and utilized in materials science.

\section{Methods}
We aim to build a publicly available knowledge base for the semantic search of experimental sections focused on Polycrystalline  materials.

\textbf{Corpus collection}: Since most scientific publications can only be accessed on specific journals, their results can not be publicly distributed, thus not satisfying our needs. We turn to the largest open-access scientific publications, S2ORC \cite{lo-etal-2020-s2orc} dataset, for acquiring all available full-text articles, specifically focusing on the subdomains of materials science and physics. However, most articles only provide abstract parts, and we obtain 838k, and 213k full text, respectively. Finally, all paragraphs are parsed by the Chemdataextractor \cite{swain2016chemdataextractor} specifically designed for the scientific domain.

\textbf{Data Filtering}: To obtain relevant information, we applied predefined key phrases (see Appendix \ref{app:kp}) suggested by materials experts to filter all relevant paragraphs from the result in the previous step, which gives us $5,846$ articles with full text. To test the recall rate of our filtering mechanism, we also test this filtering process to the full article of the test set in PcMSP \cite{yang2022pcmsp}, where we successfully retrieve 230 relevant paragraphs from 290 original examples, achieving a recall of 80\%.

\textbf{Named Entity Recognition}: To extract semantic entities within the filtered paragraphs, we utilized the Named Entity Recognition (NER) model proposed by \cite{zhong-chen-2021-frustratingly}. We follow the training setups in \cite{yang2022pcmsp} and obtain an overall F1 score of 79\% using the MatBERT trained on 50 million materials science paragraphs by \cite{walker2021impact}.

\textbf{Semantic Search}: The extracted information was then loaded into our intelligent search engine powered by Elasticsearch \footnote{https://www.elastic.co/downloads/elasticsearch}, enabling fast and flexible search capabilities. We adopt the pipeline in SynKB from \cite{bai2022synkb} for interface design.

\textbf{User Interface}: Our interface allows researchers to search for specific information, such as temperature or pressure, by entering single or multiple keywords. The system returns all relevant paragraphs, enabling quick and easy access to the most important methods in previous research.

\begin{figure*}
\centering
    \includegraphics[width=1.0\textwidth]{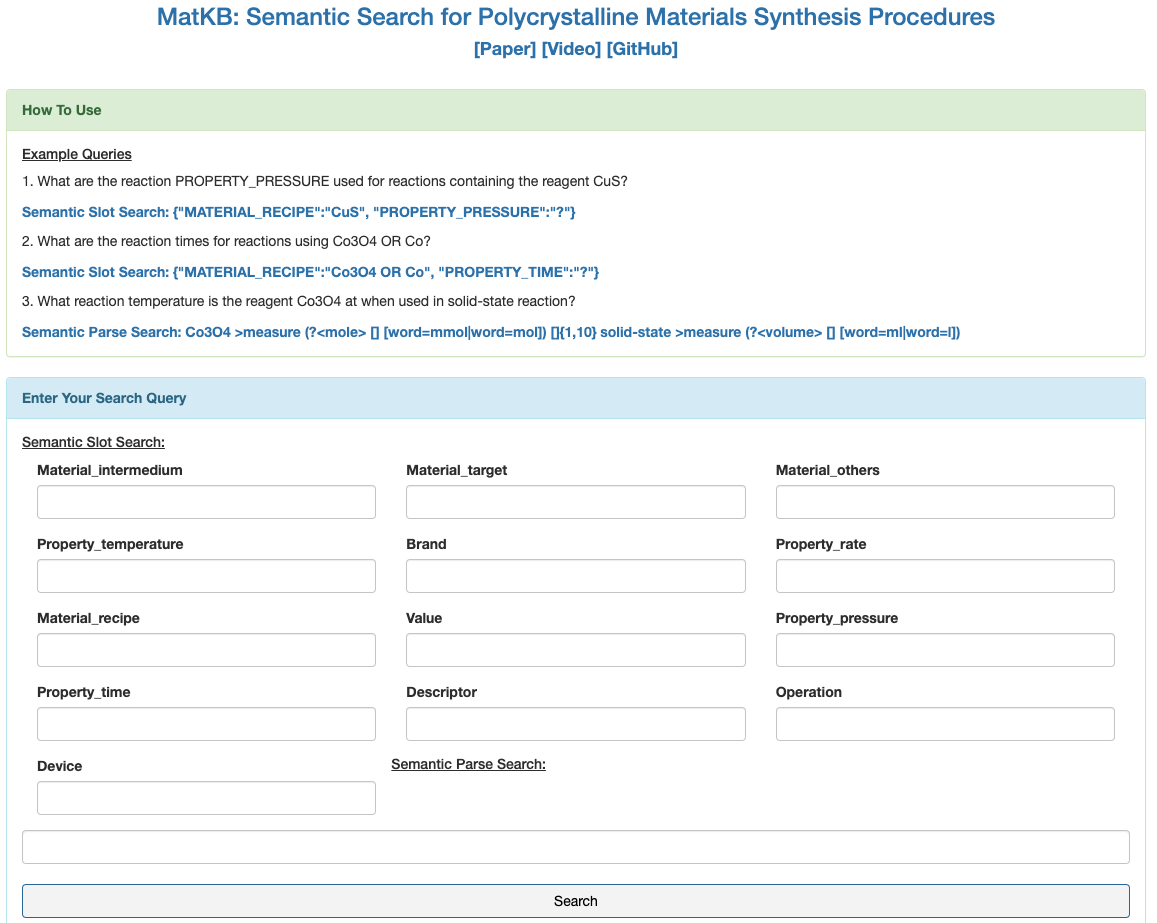} 
    \caption{\label{fig:demo} An overview of our MatKB semantic search interface. Different semantic slots can be combined or independently for search.}
\end{figure*}

\begin{figure*}
\centering
    \includegraphics[width=1.0\textwidth]{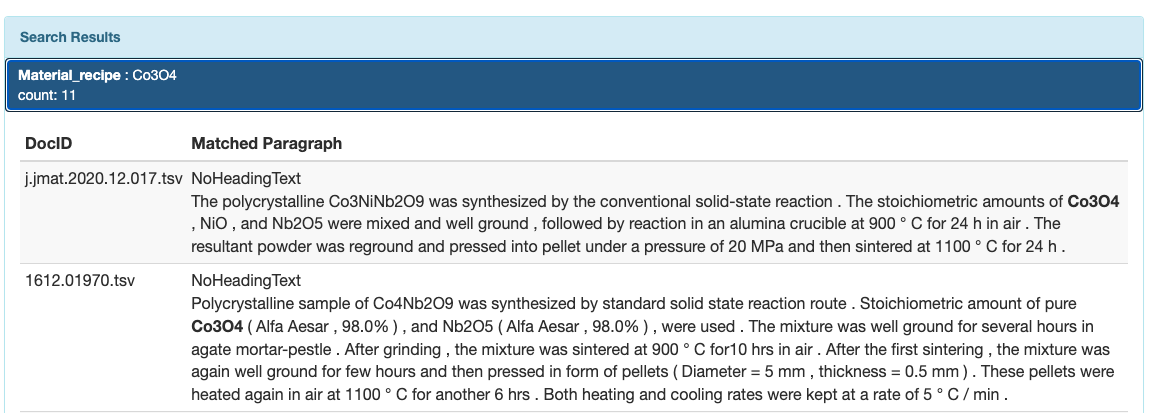} 
    \caption{\label{fig:result1} An example showing the search results by \textit{Material\_recipe:} $Co3O4$. }
\end{figure*}

\begin{figure*}
\centering
    \includegraphics[width=1.0\textwidth]{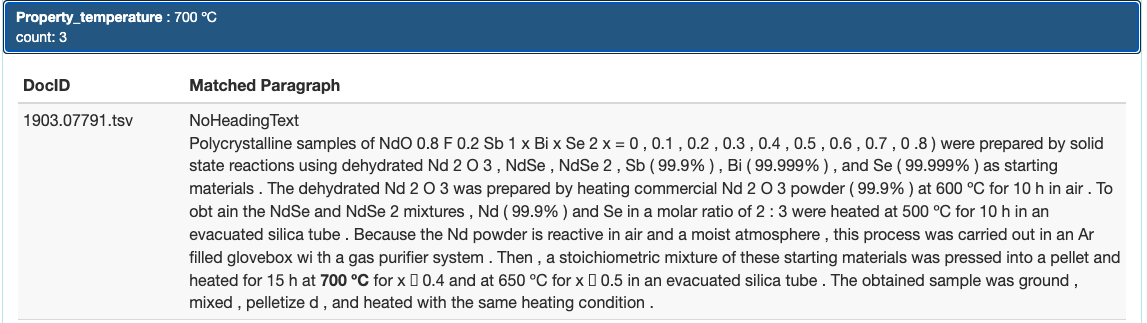} 
    \caption{\label{fig:result2} An example showing the search results by \textit{Material\_temperature:} $700$ °C.  }
\end{figure*}

\begin{figure*}
\centering
    \includegraphics[width=1.0\textwidth]{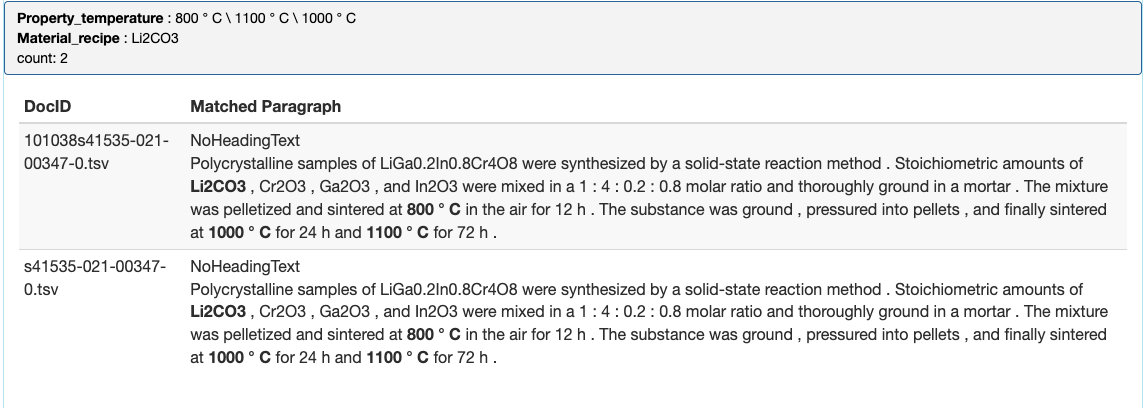} 
    \caption{\label{fig:result3} An example showing the search results by a combination of \textit{Material\_temperature:} $1000$ °C and \textit{Material\_recipe:} $Li2Co3$.  }
\end{figure*}

\section{Result}
\subsection{Statistics}
Table \ref{tab:stats} shows the statistics of predicted entity mentions in a dataset. The entity mentions are divided into 11 categories: Descriptor, Material-target, Material-intermedium, Operation, Device, Brand, Property-time, Value, Property-pressure, Material-others, Material-recipe, and Property-temperature, following the original definition in PcMSP \cite{yang2022pcmsp}. The categories are defined based on the type of information they represent. For each category, Table \ref{tab:stats} lists the number of counts (\#Count), the number of unique mentions (\#Unique), and a few examples of the mentions. The extracted dataset's total number of entity mentions is 269,808, with 29,774 unique mentions. The most frequently mentioned category is Descriptor, with 82,766 counts, followed by Operation, with 55,229 counts. The least frequently mentioned category is Property-rate, with only 2,133 counts. The information in this table provides insights into the distribution of entity mentions across the different categories, which can be useful for various data analysis and information extraction tasks.

\begin{table*}[h]
\centering
\scalebox{.90}
{
\begin{tabular}{c|c|c|c }
\hline
\textbf{Name} & \textbf{ \#Count } &\textbf{ \#Unique } & \textbf{Examples} \\
%\toprule
\hline
\textit{Descriptor} & 82,766 & 7,721 & polycrystalline, different, powder, single \\
\textit{Material-target}  & 11,651 & 1,063 & SiC, FeSe, ZnO, LaFeAsO  \\
\textit{Material-intermedium}  & 18,956&1,356  & solution, grains, powders, pellets  \\
\textit{Operation}    & 55,229& 3,993 & added, arc, heat, grinding \\
\textit{Device}    & 15,659&2,163 & tube, furnace, ampoule, crucible \\
\textit{Brand} & 5,241&1,671 & Sigma-Aldrich, Rigaku, Hitachi, Bruker \\
\textit{Property-time}  & 5,103&794 & 24 h, 30 min, 3 h, 1.5 hours \\
\textit{Value}  & 24,045&3,295 & 10 mg, stoichiometric amounts, 2 ml, around 3 g  \\
\textit{Property-pressure}  & 9,466&2,190 & nitrogen, ambient pressure, air, 20 KPa  \\
\textit{Material-others}  & 8,294&1,338 & ethanol, water, carbon, silicon  \\
\textit{Material-recipe}  & 18,341&1,218 & Al, Si, Ga, Zn \\
\textit{Property-temperature}  & 12,924&2,303 & room temperature, 1000 °C, below 600 °C, about 100 °C  \\
\textit{Property-rate}  & 2,133&669 & cooling rate, 1 K/min, approximately 2 K/min, air \\
\hline
\textit{Total}  & 269,808 & 29,774 &  \\
\hline
\end{tabular}
}
\caption{Predicted entity mention statistics and corresponding examples.}
\label{tab:stats}
\end{table*}

\subsection{Search}
In Figure \ref{fig:demo}, we show an overview of our search interface, where we can perform a search according to our predefined semantic slots. For example, the results in Fig. \ref{fig:result1} are obtained by a slot search of \textit{Material\_recipe:} $Co3O4$. Besides, we additionally show more search examples in Fig. \ref{fig:result2} and \ref{fig:result3}. Compared with traditional search engines like Google or scholar search platform like Google scholar or Semantic Scholar, our pre-extracted entities can return us with precise experimental sections without further click-into publishers' websites and do tediously manual filtering. We hope such a tool can help materials scientists save time looking for correct references for experiments. Furthermore, since we return multiple results with different experimental procedures, material scientists can also compare the differences between those methods for designing their experiments.

\section{Conclusion}
In conclusion, we have presented a new approach for extracting structured knowledge from large amounts of research articles in materials science. Our method leverages NLP techniques to identify entities and experimental sections and builds an extensive knowledge base for easy search and retrieval. The proposed system demonstrates superiority over traditional search methods like Google by instantly returning experimental sections based on specific entity queries. Our results show that our approach can effectively extract valuable information and provide a comprehensive overview of current research in the field of materials science.

Future work will focus on expanding our knowledge base to cover a broader range of research articles and improving the accuracy of our entity recognition and experimental section extraction models. Additionally, we plan to enhance the user experience of the search website by incorporating interactive visualizations and more advanced search algorithms. We believe that this system has the potential to greatly improve the efficiency and effectiveness of research in the field of materials science and ultimately contribute to scientific advancements in this area.

\section{Limitations}

Data Bias: The study's results may be biased by the limited scope of the data used, and any biases present within the data.

Model Limitations: The results are only as reliable as the Named Entity Recognition (NER) model used, and any limitations or inaccuracies in the model could affect the results.

User Error: The accuracy of the results may be impacted by user error, such as incorrect keywords or misunderstandings of the system functionality.

It is important to carefully consider and address these potential risks in the study's design and implementation to ensure the results' validity and reliability.

\subsubsection*{Acknowledgments}
We gratefully acknowledge support from the UC Santa Barbara NSF Quantum Foundry funded via the Q-AMASEi program under NSF award DMR-1906325. Any opinion or conclusions expressed in this material are those of the author(s) and do not necessarily reflect the views of the National Science Foundation.

\bibliography{iclr2023_conference}
\bibliographystyle{iclr2023_conference}

\section{Appendix}
%You may include other additional sections here.
\subsection{Key Phases}\label{app:kp}

\textbf{Strategy 1}: key\_list = [ 'powder samples were prepared', 'powders were obtained', 'Polycrystalline ingots', 'ground together and pressed into pellets', 
    'starting materials were ground together', 'were prepared using bulk solid state methods', 'arc-melting stoichiometric quantities',
     'ground together and pressed into pellets', 'starting materials were ground together', 'polycrystalline/Polycrystalline samples were', 'polycrystalline/Polycrystalline sample was' ]

\textbf{Strategy 2}: First it satisfies that 'polycrystalline' and 'Polycrystalline' in text and then perform a second round filtering, 
key\_list = ['were/was synthesized/prepared', 'were/was first synthesized/prepared', 'were/was used', 'were/was first used', 'were/was obtained', 'were/was first obtained', 'were/was achieved', 'were/was first achieved']

\end{document}

%% file: math_commands.tex
\usepackage{amsmath,amsfonts,bm}

\def\eqref#1{equation~\ref{#1}}
\def\1{\bm{1}}

\DeclareMathAlphabet{\mathsfit}{\encodingdefault}{\sfdefault}{m}{sl}
\SetMathAlphabet{\mathsfit}{bold}{\encodingdefault}{\sfdefault}{bx}{n}